\documentclass[runningheads]{llncs}
\usepackage{graphicx}

\usepackage{tikz}
\usepackage{comment}
\usepackage{amsmath,amssymb} 
\usepackage{color}
\usepackage{cite} 
\usepackage{mathrsfs}
\usepackage{booktabs}

\usepackage[accsupp]{axessibility}  

\begin{document}
\pagestyle{headings}
\mainmatter

\title{Multi-Task Learning for Emotion Descriptors Estimation at the fourth ABAW Challenge} 

\author{Yanan Chang \and Yi Wu \and Xiangyu Miao \and Jiahe Wang \and Shangfei Wang} 
\institute{University of Science and Technology of China, Hefei, China \\
\email{\{cyn123, wy221711, mxy3369, pia317\}@mail.ustc.edu.cn}\\
\email{sfwang@ustc.edu.cn}}

\maketitle

\begin{abstract}

Facial valence/arousal, expression and action unit are related tasks in facial affective analysis. However, the tasks only have limited performance in the wild due to the various collected conditions. The 4th competition on affective behavior analysis in the wild~(ABAW) provided images with valence/arousal, expression and action unit labels. In this paper, we introduce multi-task learning framework to enhance the performance of three related tasks in the wild. Feature sharing and label fusion are used to utilize their relations. We conduct experiments on the provided training and validating data.

\end{abstract}

\section{Introduction}
Facial valence/arousal prediction, facial expression recognition, and facial action unit~(AU) recognition are three significant tasks for facial affective analysis. The three tasks are related. Leveraging their relationships will benefit for their performance. However, it is challenging to conduct the three tasks in the wild due to the complex collected environment.

The 4th competition on affective behavior analysis in the wild~(ABAW) has provided s-Aff-Wild2 database~\cite{kollias2022abaw,kollias2022abaw, kollias2021distribution, kollias2021affect, kollias2020deep, kollias2020va, kollias2019expression, kollias2019deep, kollias2018photorealistic, zafeiriou2017aff, kollias2017recognition} for the first challenge, the multi-task learning~(MTL) challenge. Images with valence/arousal values, expressions and AUs are provided for the training and validation. The key is how to leverage the correlations among valence/arousal, AU and expression tasks to improve the performance in the wild.

In order to enhance the performance of three tasks in the wild, we introduce multi-task learning framework based on feature sharing and label fusion. One multi-task learning component based on shared features is used to perform three tasks jointly. Considering the difference of tasks, we also perform each task by single component. The predictions from the multi-task component and the single tasks are fused as the final results.


\section{Method}

    \begin{figure*}[htbp]
        \centering
        \includegraphics[scale=0.46]{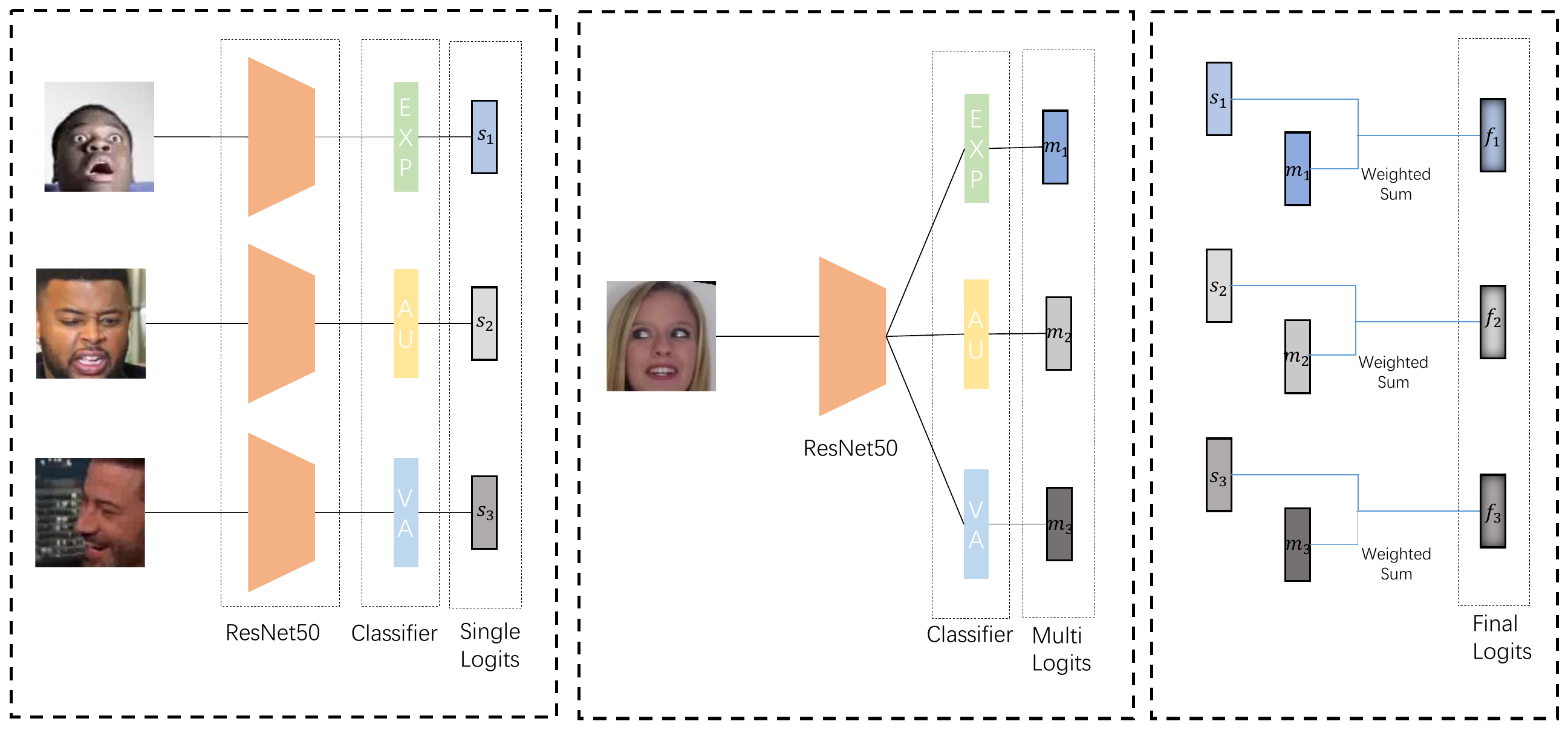}
        \caption{Overview of our proposed framework}
    \end{figure*}

    Figure-1 illustrates the framework of our proposed approach. The proposed method combines Single-Task Learning and Multi-Task Learning, and processes the results obtained by both methods through a weighted manner.

    Specifically, in Single-Task Learning, we use Resnet50 as the feature extractor for Expression Recognition, AU Recognition, and VA Regression. The corresponding classifier maps high-dimensional features to low-dimensional labels respectively.

    Considering that the above three tasks are related to facial motions, we adopt the popular hard parameter sharing method in multi-task. The three tasks share the same Resnet50, and simultaneously map the features extracted by the network to three low-dimensional labels.

    It is verified by experiments that the performance of Multi-Task Learning is better than that of Single-Task Learning on VA Regression, and worse on Expression Recognition and AU Recognition. The intuitive idea is to integrate the prediction results of the two methods, and the combined weight is needed to adjust manually.

    Through the statistics of various expressions and various AU samples in the training dataset, we found that there is a serious data imbalance problem in the dataset.

    To mitigate the problem, for Expression Recognition, we employ a Weighted Cross-Entropy Loss Function to weight samples of various expressions. Specifically, the weight of various samples is the quotient of the total number of samples and the total number of various samples. Let E and $f_{exp}$ be the Resnet50 encoder and the expression classifier, for an input image x and its corresponding expression label $y_{exp}$, the weighted Cross-Entropy Loss Function is formulated as:

   $$\mathcal{L}_1=w_y*CE(f_{exp}(E(x)),y_{exp})$$

   where CE represents Cross-Entropy and $w_y$ represents the weight of the y-th expression.

    For AU Recognition, we utilize the recently proposed Focal Loss to weight each sample. The core of the loss is that, let the predicted probability be p, for positive samples, its weight is $1 - p$, and for negative samples, its weight is p. Let $f_{au}$ and $y_{au}$ be the au classifier and au labels, the Focal Loss Function is formulated as:

    $$\mathcal{L}_2=FL(f_{au}(E(x)),y_{au})$$

    where FL represents Focal Loss.

    For VA Regression, in order to maximize the Concordance Correlation Coefficient (CCC) metric, we minimize $1-CCC$. Let $f_{va}$ and $y_{va}$ be the va regressor and va labels, the Loss Function for va is formulated as:

    $$\mathcal{L}_3=1-CCC(f_{va}(E(x)),y_{va})$$

    The total loss is defined as:

    $$\mathcal{L}=\mathcal{L}_1+\mathcal{L}_2+\mathcal{L}_3$$

    During the inference phase, the final prediction is composed of two results from Single-Task Learning and Multi-Task Learning respectively, which is defined as:

    $$p_{final}=\lambda*p_{single}+(1-\lambda)*p_{multi}$$

    where $\lambda$ is a weight factor which connects the results obtained from two learning methods, and for VA Regression, $\lambda$ is set to 0.4, for Expression and AU recognition, $\lambda$ is set to 0.6.

\section{Experiments}
In this section, we conduct experiments on the s-Aff-Wild2 database~\cite{kollias2022abaw}.

\subsection{Experimental database}
The s-Aff-Wild2 database is selected from the Aff-Wild2 databases~\cite{kollias2022abaw, kollias2021distribution, kollias2021affect, kollias2020deep, kollias2020va, kollias2019expression, kollias2019deep, kollias2018photorealistic, zafeiriou2017aff, kollias2017recognition}. There are 142382 images for training and 26876 images for validation. For each image, valence and arousal values, expressions, and action units are labeled. Specifically, the values of valence and arousal are ranged from -1 to 1. Eight kinds of expressions are labeled, including six basic expressions, neutral, and other. For AU recognition, 12 AUs~(1,2,4,6,7,10,12,15,23,24,25,26) are annotated as active~(1) or inactive~(0). In addition, there are some annotation values which should not be considered, for example -1 for AUs and expressions, and -5 for valence and arousal. For our training and validation, we do not use the annotation values.

\subsection{Implementation Details}
The feature extraction network is based on ResNet-50~\cite{He2015}. The predictors for each task include one linear layer. The output size for valence and arousal is 2. The output sizes for AU recognition and expression recognition are 12 and 8, respectively. The overall framework is implemented by PyTorch~\cite{paszke2019pytorch}. Adam optimizer~\cite{kingma2014adam} is used to optimize the network. The batch size and learning rate are set to 64 and 5e-5, respectively.
In order to solve the unbalanced data distribution, we leverage different class weights for facial expression recognition. The weights are calculated by counting the numbers of different class images. The class with less images will have higher weight.

\subsection{Evaluation Metrics}
According to the performance assessment rules of the competition~\cite{kollias2022abaw}, we evaluate the performance of valence and arousal task by the mean Concordance Correlation Coefficient (CCC). The AU recognition and expression recognition are evaluated by the average F1 score. The performances of three tasks are summed as the final evaluation index, which is formulated as:

$$P=0.5*(CCC\_arousal+CCC\_valence)+0.125*\sum{F1\_expr}+\sum{F1\_au/12}$$

\subsection{Experimental Results}

When training a separate model for each task, that is, single-task learning. The performance of each task and the sum of the performance of all tasks are shown in Table \ref{table1}.

\begin{table}[!htbp]
    \centering
    \caption{The performance on s-Aff-Wild2 database using single-task learning}
    \label{table1}
    \begin{tabular}{c|c|c|c|c}
        \hline
               &   VA  &  expr &   AU  &   ALL\\
        \hline
        metric & 29.36 & 24.66 & 47.67 & 101.69\\
        \hline
    \end{tabular}
\end{table}

We also make the three tasks share a feature extraction network, and each task has its own classifier, that is, multi-task learning. Similarly, the performances are shown in Table \ref{table2}.

\begin{table}[!htbp]
    \centering
    \caption{The performance on s-Aff-Wild2 database using multi-task learning}
    \label{table2}
    \begin{tabular}{c|c|c|c|c}
        \hline
               &   VA  &  expr &   AU  &   ALL\\
        \hline
        metric & 36.57 & 22.80 & 41.02 & 100.39\\
        \hline
    \end{tabular}
\end{table}

\section{Conclusions}

We utilize multi-task learning method for the MTL challenge of 4th ABAW competition. The relationships among the related multiple tasks are leveraged by sharing feature and fusing the final predictions. Experimental evaluations are conducted on the database provided from the challenge.



%
%
\bibliographystyle{splncs04}
\bibliography{egbib}
\end{document}